\documentclass[letterpaper]{article} 
\usepackage{aaai25}  
\usepackage{times}  
\usepackage{helvet}  
\usepackage{courier}  
\usepackage[hyphens]{url}  
\usepackage{graphicx} 
\usepackage{natbib}  
\usepackage{caption} 
\usepackage{algorithm}
\usepackage{algorithmic}
\usepackage{amsmath}
\usepackage{booktabs}
\usepackage{array}
\usepackage{xcolor} 
\usepackage{mdframed} 

\definecolor{softblue}{rgb}{0.2, 0.4, 0.6}   
\definecolor{reddishbrown}{rgb}{0.8, 0.3, 0.2} 
\definecolor{mygray}{gray}{0.9}  
\definecolor{deepblue}{rgb}{0.1, 0.3, 0.6}   

\usepackage{newfloat}
\usepackage{listings}
\DeclareCaptionStyle{ruled}{labelfont=normalfont,labelsep=colon,strut=off} 
\floatstyle{ruled}
\newfloat{listing}{tb}{lst}{}
\floatname{listing}{Listing}
\title{Efficient Language-instructed Skill Acquisition via Reward-Policy Co-Evolution}

\author{
    Changxin Huang\textsuperscript{\rm 1},
    Yanbin Chang\textsuperscript{\rm 1},
    Junfan Lin\textsuperscript{\rm 2}\thanks{Corresponding author},
    Junyang Liang\textsuperscript{\rm 1},
    Runhao Zeng\textsuperscript{\rm 3},
    Jianqiang Li\textsuperscript{\rm 1}\footnotemark[1]}

\affiliations{
    \textsuperscript{\rm 1}National Engineering Laboratory for Big Data System Computing Technology, Shenzhen University, Shenzhen, China\\

    \textsuperscript{\rm 2}Peng Cheng Laboratory, Shenzhen, China\\
    
    \textsuperscript{\rm 3}Artificial Intelligence Research Institute, Shenzhen MSU-BIT University, Shenzhen, China\\
    {}
    huangchx@szu.edu.cn, changyanbin2023@email.szu.edu.cn, linjf@pcl.ac.cn, \\ liangjunyang2018@email.szu.edu.cn, zengrh@smbu.edu.cn, lijq@szu.edu.cn
}

\usepackage{bibentry}

\begin{document}

\maketitle

\begin{abstract}
The ability to autonomously explore and resolve tasks with minimal human guidance is crucial for the self-development of embodied intelligence. 
Although reinforcement learning methods can largely ease human effort, it's challenging to design reward functions for real-world tasks, especially for high-dimensional robotic control, due to complex relationships among joints and tasks. Recent advancements large language models (LLMs) enable automatic reward function design. 
However, approaches evaluate reward functions by re-training policies from scratch placing an undue burden on the reward function, expecting it to be effective throughout the whole policy improvement process. 
We argue for a more practical strategy in robotic autonomy, focusing on refining existing policies with policy-dependent reward functions rather than a universal one.
To this end, we propose a novel reward-policy co-evolution framework where the reward function and the learned policy benefit from each other's progressive on-the-fly improvements, resulting in more efficient and higher-performing skill acquisition. 
Specifically, the reward evolution process translates the robot's previous best reward function, descriptions of tasks and environment into text inputs. These inputs are used to query LLMs to generate a dynamic amount of reward function candidates, ensuring continuous improvement at each round of evolution. For policy evolution, our method generates new policy populations by hybridizing historically optimal and random policies. Through an improved Bayesian optimization, our approach efficiently and robustly identifies the most capable and plastic reward-policy combination, which then proceeds to the next round of co-evolution.
Despite using less data, our approach demonstrates an average normalized improvement of 95.3\% across various high-dimensional robotic skill learning tasks.

\end{abstract}
\section{Introduction}

\begin{figure}[t]
    \centering
    \includegraphics[width=1.0\columnwidth, trim=10 380 510 10, clip]{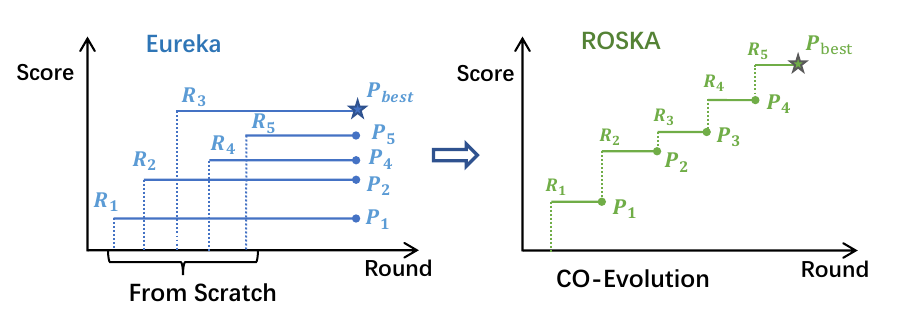}
    \caption{Comparison of main differences between our method and Eureka.}
    \label{fig:Comparison}
\end{figure} 
As affordable robots are increasingly common, it becomes more and more crucial to develop algorithms that enable robots to understand commands, solve problems, and self-evolve, reducing the need for constant human supervision. Reinforcement learning (RL), which employs reward functions to decrease reliance on human supervision \cite{haarnoja2024learning}, has been a triumph across various domains, including video games \cite{alonso2020deep} and the strategic game of chess \cite{silver2018general}. Yet, when we shift our focus to the intricate realm of high-dimensional robot locomotion \cite{radosavovic2024real}, the task of formulating a reward function for fundamental movement becomes notably complex. It's not just a matter of technical know-how; it's about navigating the unique intricacies of each robot's form and joint configuration \cite{andrychowicz2020learning}. 

In recent years, with the rapid growth of large language models (LLMs) \cite{DBLP:journals/corr/abs-2402-06196}, numerous approaches have been proposed to enhance robots' ability to execute tasks based on human instructions. For instance, Google's SayCan approach integrates LLMs with robotic affordances, allowing robots to generate feasible task plans grounded in real-world actions \cite{brohan2023can}. Similarly, the Code as Policies method leverages LLMs to autonomously generate robot policy code from natural language commands, enabling robots to generalize to new instructions and environments without additional training \cite{liang2023code}.  Among these, Euraka is a standout initiative, marking a pioneering step in the self-design of reward functions tailored to task instructions with the help of LLMs \cite{DBLP:conf/iclr/MaLWHBJZFA24}. They aim to develop a universally applicable reward function that can guide a policy with randomly initialized parameters to learn and perform tasks directed by language instructions.

Despite its success, Eureka is still confined by the traditional RL framework, which relies on one reward function to provide feedback throughout the policy improvement process. This can lead to inefficient and ineffective policy optimization. Firstly, each intermediate reward function necessitates learning a policy from scratch, as shown on the left of Fig.\ref{fig:Comparison}. Secondly, finding a universal reward function is non-trivial because it must be sufficiently comprehensive to consistently offer task-relevant feedback for different state-action-state transitions across the whole policy improvement process. This not only results in an enormous search space but also demands a more sophisticated prompt design for the LLM as the complexity of tasks increases. This orthogonal approach to improving reward functions and policies separately is costly and can impede real-world applications.

In our exploration, we've identified a transformative opportunity to refine the traditional RL model, gearing it towards greater efficiency and real-world practicality in the era of LLM. Recognizing the prowess of LLMs to craft reward functions for specific commands, we propose to harness this further: we aim to empower LLMs to autonomously adapt and refine the reward functions in conjunction with policy improvements. In this paper, we propose a novel \textbf{reward-policy co-evolution framework} for efficient language-instructed \underline{RO}bot \underline{SK}ill \underline{A}cquisition (\textbf{ROSKA}), where the reward function and the policy co-evolve in tandem, rather than separately, exponentially speeding up the learning process, as sketched in the right of Fig.\ref{fig:Comparison}.

Specifically, for the evolution of the reward function, we've crafted a cutting-edge LLM-driven mechanism for evolving reward functions, ensuring steady enhancement at each evolutionary phase. This is accomplished by dynamically expanding the population of reward function candidates generated by the LLM, using the historically top-performing reward function as a benchmark for this expansion. 
In regards to the policy evolution, to adapt the ongoing policy to become both capable and plastic to the new reward functions, we initiate with policy candidates by fusing the parameters of the previous best policy with a dash of randomized parameters in varying proportions. The top-performing fused policy will be selected for further optimization under their respective reward functions. 
To quickly identify the promising fused policies from countless possible fused policies, we've implemented Short-Cut Bayesian Optimization (SC-BO) for a faster policy evolution. SC-BO leverages the observation that different fused policies generally diverge after a few updates to conduct an early stop optimization. Additionally, SC-BO only uses a limited number of search points allowing for fast searches, and the dynamic population of the reward evolution will guarantee the continual refinement of policies. This method achieves superior results with fewer iterations compared to vanilla BO.

At the end of each reward-policy co-evolution cycle, the most effective reward-policy combination will initiate the subsequent round of reward-policy co-evolution.

Extensive experimental results demonstrate that our approach utilizes only 89\% of the data and achieves an average normalized improvement of 95.3\% across various high-dimensional robotic skill-learning tasks, highlighting its effectiveness in enhancing the adaptability and precision of robots in complex environments.

\section{Related Work}

\textbf{Designing reward functions for robot.} Designing reward functions for applying reinforcement learning to robotic tasks has long been a significant challenge. Existing approaches can be broadly categorized into manually designed rewards  and automated reward generation \cite{DBLP:conf/corl/0003GFKLACEHHIX23}. Manually designed rewards rely heavily on extensive domain knowledge and experience \cite{booth2023perils}. 

Automated reward design includes Inverse Reinforcement Learning (IRL) \cite{pinto2017robust} and LLM-based reward generation \cite{DBLP:conf/iclr/MaLWHBJZFA24}. IRL is a data-driven approach that derives reward functions from demonstration data \cite{ziebart2008maximum}. 
However, IRL relies on high-quality demonstration data, which is often expensive to collect, particularly for robotic applications.

Recently, large language models (LLMs) have been employed to design reward functions by directly converting natural language instructions into rewards  \cite{lin2022inferring, hu2023language}, such as in "text2reward" \cite{DBLP:journals/corr/abs-2309-11489} and "language2reward" \cite{DBLP:conf/corl/0003GFKLACEHHIX23}. However, these methods typically require predefined reward templates or initial reward functions. To design reward functions from scratch, NVIDIA researchers introduced the Eureka framework, which employs a multi-round iterative process to design reward functions and uses RL to train robots for complex skills \cite{DBLP:conf/iclr/MaLWHBJZFA24}. The reward functions generated by Eureka must be trained to verify their effectiveness, with results fed back to the LLM for further refinement, leading to high training costs. In contrast, our approach fine-tunes a pre-trained policy on a new reward function, significantly enhancing data efficiency.

\textbf{Leveraging pre-trained policies to enhance training efficiency in RL.} The utilization of pre-trained policies to fine-tune models in new environments, thereby improving training efficiency, has proven effective in robotic tasks \cite{DBLP:journals/corr/abs-2210-05178, walke2023don}. Common approaches include offline RL \cite{kumar2020conservative} and meta RL \cite{wang2023rapid}. Offline RL algorithms develop robot control policies from pre-existing demonstration data or offline interaction datasets. These pre-trained policies can then be utilized during the online fine-tuning phase to adapt to novel tasks \cite{lee2022offline}. Meta RL focuses on training policies across diverse tasks that can rapidly adapt to new ones \cite{arndt2020meta}. We propose incorporating pre-trained policies into the LLM-based reward design process to accelerate training, avoiding the need to start from scratch. To align the pre-trained policy with the designed reward, we introduce a novel policy evolution method using Bayesian Optimization to determine the optimal inheritance ratio.

\begin{figure*}[t]
    \centering
    \includegraphics[width=0.90\textwidth, trim=115 65 110 65, clip]{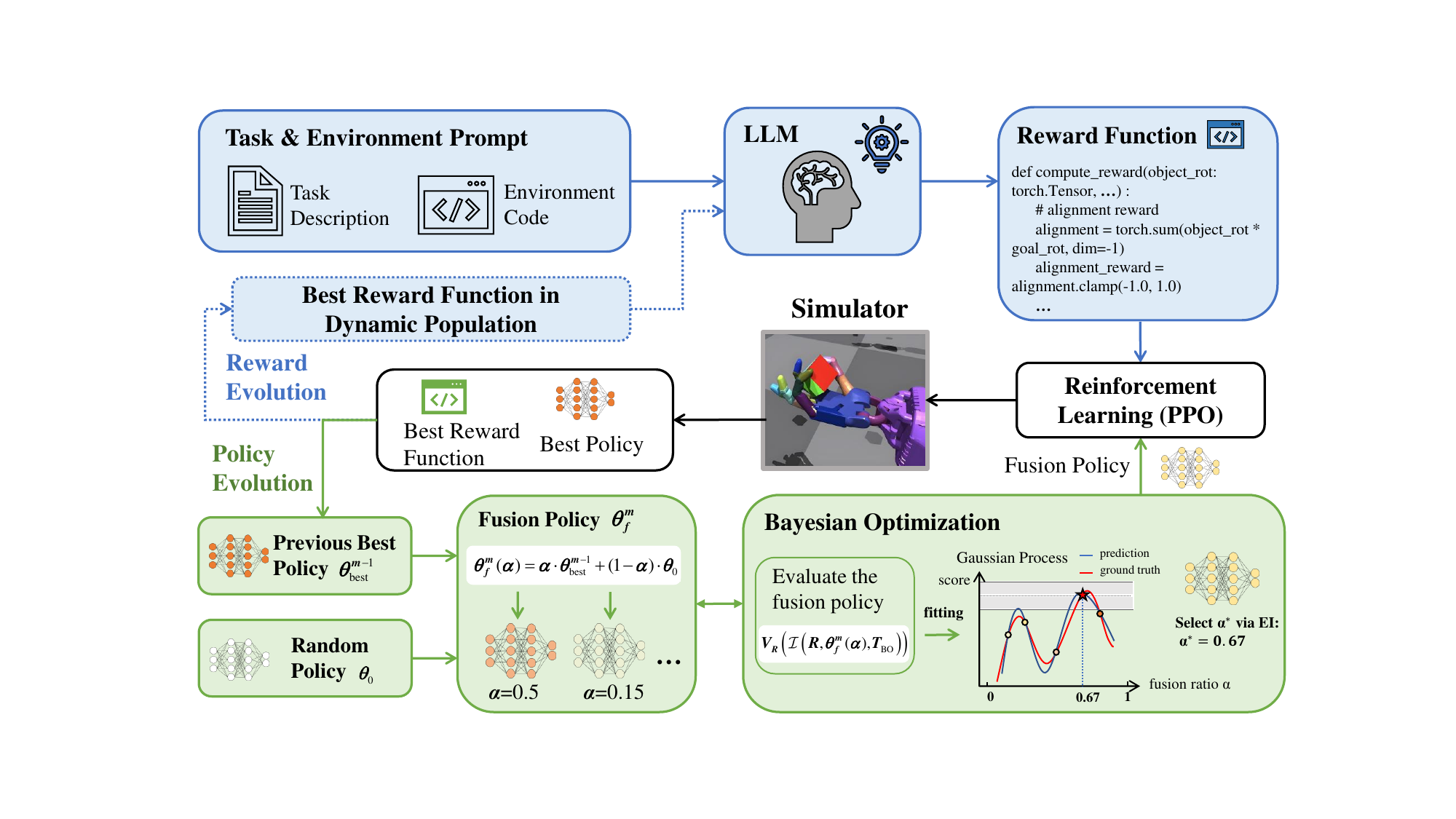}
    \caption{Overview of the proposed reward-policy co-evolutionary framework, illustrating the iterative refinement of reward functions and policies through mutual feedback between a large language model (LLM), reinforcement learning (PPO), and Bayesian optimization, enabling efficient and effective skill acquisition.}
    \label{fig:framework}
\end{figure*}

\section{Preliminary} \label{sec:preliminary}

\textbf{Reinforcement learning in robotic skill learning.} Multi-joint robotic skill acquisition can be formulated as a Markov Decision Process (MDP) \cite{puterman1990markov}, where the robot interacts with the environment $E$. An MDP is represented as $(S, A, P, R, \gamma)$, which includes the state space $S$, action space $A$, state transition probability function $P$, reward function $R$, and discount factor $\gamma \in (0, 1]$. At each time step $t$, the robot observes a state $s_t \in S$ and selects an action $a_t$ according to the policy $\pi(s_t)$. The environment then transitions to the next state $s_{t+1} \sim P(s_{t+1} \mid s_t, a_t)$, and the robot receives a reward $r_t = R(s_t, a_t, s_{t+1})$ and updates its state. The return for state $s_t$ is defined as the cumulative $\gamma$-discounted reward: $\sum_{i=t}^T \gamma^{i-t} r_i$. Reinforcement learning aims to optimize the policy $\pi$ by maximizing the expected return from the initial state. Formally, we define $\theta_p$ as the parameters of policy $\pi$ after $p$ updates, and
\begin{align}
    \theta_{p}=\mathcal{I}(R, \theta_0, p), \label{equ:policy_improve}
\end{align}
where $\theta_0$ represents the randomly-initialized parameters of policy $\pi$, and function $\mathcal{I}(R, \theta, p)$ stands for policy improvement process that the parameters $\theta$ are improved after $p$ updates given the reward function $R$. To this end, the optimal policy parameters can be formulated as:
\begin{align}
    \theta^* = \mathcal{I}(R, \theta_0, \infty).
\end{align}
  
\textbf{language-instructed reward function generation.} The design of the reward function is crucial in reinforcement learning and often relies on the experience of researchers and practitioners, involving iterative adjustments through trial and error. This is particularly challenging for multi-joint robotic skill acquisition, as it requires considering multiple factors, such as robot stability and task relevancy.
To address the challenge of designing complex reward functions, a recent study, i.e., Eureka~\cite{DBLP:conf/iclr/MaLWHBJZFA24}, utilizes Large Language Models (LLMs) to generate executable reward function code progressively. Specifically, given the task description $I_d$ and the environment code $I_e$, the LLM generate reward functions by: 
\begin{equation}\label{eq:llm_base}
\mathbf{R}^n = \text{LLM}(I_d, I_e, R^{n-1}_\text{best}, V(\theta)),
\end{equation}
where $\mathbf{R}^n = [R^n_1, R^n_2, ..., R^n_K]$ represents the set of $K$ reward functions generated by the LLM at round $n$, $V(\theta)$ stands for the return of the policy $\pi_\theta$ under the ground-truth sparse reward function $R^*$ and $R^{n-1}_{best} \in \mathbf{R}^{n-1}$ is the most suitable reward code in last round:
\begin{align}
    R_\text{best}^{n-1} = \max_{R^{n-1}_k \in \mathbf{R}^{n-1}} V\left( \mathcal{I}(R^{n-1}_k, \theta_0, T_\text{max})\right). \label{equ:eureka}
\end{align}
Eq.(\ref{equ:eureka}) means that each generated reward functions are used to optimize a policy from scratch $\theta_0$, and after $T_\text{max}$ training epochs of improvement, the reward function whose policy obtains the best performance of the task is considered the most suitable reward function at this round. 

\section{Method} \label{sec:method}
\subsection{Overview} \label{sec:Overview}
In this paper, we propose a novel reward-policy co-evolution framework for \underline{RO}bot \underline{SK}ill \underline{A}cquisition (\textbf{ROSKA}), that enables robots to learn how to complete language-instructed tasks via automatic reward functions and policy co-evolution, significantly improving the efficiency and efficacy. As illustrated in Fig.\ref{fig:framework}, ROSKA can be roughly divided into two distinctive yet mutually enhanced modules: \textbf{reward evolution} and \textbf{policy evolution}.

Briefly, the reward evolution module (as shown in the upper part of Fig. \ref{fig:framework} prompts the LLM with the robot's task and environment descriptions to generate a dynamic set of reward functions. The most effective reward (evaluated after policy evolution) from this set is selected to proceed to the next evolution cycle; as for policy evolution (lower part in Fig. \ref{fig:framework}, given a newly generated reward function, rather than starting from scratch, ROSKA builds on the previous round's best-performed policy to leverage its learned capability, blending its parameters with random noise to maintain plasticity. We use Bayesian Optimization with an early stop mechanism to find the optimal blending ratio that can balance between retaining learned skills and allowing for new learning. The policy with the best performance across all candidate rewards will proceed to the next round the policy evolution.

In the following sections, we will elaborate on each of these modules in detail.

\subsection{Reward Evolution with Dynamic Population} \label{sec:Reward Evolution}

In Eureka~\cite{DBLP:conf/iclr/MaLWHBJZFA24}, within each round of reward searching, a set number of $K$ reward functions are generated by a Large Language Model (LLM), as depicted in Eq.(\ref{eq:llm_base}). Each reward function is then thoroughly tested by training a new policy, as shown in Eq.(\ref{equ:eureka}). However, we've discovered a potential performance degradation with this approach: if the size of candidate reward functions is small, there is a chance that no superior reward functions could be identified. Using a large size could mitigate this, but it could be costly. After all, interacting with an LLM isn't cheap, and only one reward function will be chosen, rendering the rest obsolete.

To address this, we have introduced an approach to dynamically adjust the size of the population of the reward functions $\mathbf{R}_\text{DP}$ in our reward evolution process. This allows the size of the population to increase when a larger exploration is needed to witness an improvement. Formally,
\begin{equation}\label{eq:llm_rew}
\mathbf{R}_\text{DP}^{m} = \text{LLM}\left(I_d, I_e, R_\text{best}^{m-1}, V(\theta)\right),
\end{equation}
where $\mathbf{R}_\text{DP}^{m} = [R^m_1, R^m_2, ..., R^m_\text{DP}]$. We use $m$ to indicate the $m$-th DP-round to distinguish it from the concept of round mentioned previously.

To determine when to increase the size of $\mathbf{R}_\text{DP}^{m}$, analogical to how world records in the Olympics inspire athletes to push their limits and excel, we use the top-performing reward function from the previous DP-round, i.e., $R_\text{best}^{m-1}$ as a benchmark, and repeating the process of Eq.(\ref{eq:llm_rew}) to generate $K$ reward functions each time until a reward function with better performance than $R_\text{best}^{m-1}$ is witnessed.  Interestingly, we found that with the same total amount of generated reward functions, our method dynamically allocates the number of query rounds at each DP-round, and can achieve a consistent improvement upon using the fixed size of candidates at each round. This necessitates the dynamic sizes during reward evolution to avoid performance degradation.

And to effectively and efficiently identify the best reward function from $\mathbf{R}_\text{DP}^{m}$, instead of evaluating each reward function by training a policy from scratch like Eq.(\ref{equ:eureka}), we propose to evaluate them based the evolved policy which is modified from the best-performed policy with parameters $\theta_\text{best}^{m-1}$ of the previous DP-round.
\begin{align}
    R_\text{best}^{m} = \max_{R_k^{m} \in \mathbf{R}_\text{DP}^{m}} V\left( \mathcal{I}_\text{evolve}\left(R^{m}_k, \theta_\text{best}^{m-1}, T_\text{max}\right)\right), \label{equ:reward_evolution}
\end{align}
where $\mathcal{I}_\text{evolve}$ is the policy evolving process elaborated in the following section.

\subsection{Policy Evolution via Bayesian Optimization}
\label{sec:Policy Evolution}
To fully leverage the knowledge from previously trained policies and enhance training efficiency, each DP-round reuses the parameters of the best-performed policy parameters from the previous DP-round. This means the policy does not learn from scratch given a new reward function, unlike traditional policy improvement formulated in Eq.(\ref{equ:policy_improve}) adopted by Eureka in Eq. (\ref{equ:eureka}), largely eliminates the initial learning process where a policy is not able to perform basic operations, significantly improving the sample efficiency of RL training.

Unfortunately, when LLM designs a reward function that's quite different from the previous one, simply copying the previous policy parameters could result in a policy with a slow convergence on the new reward function, or it might not converge at all, as noted in \cite{DBLP:journals/corr/ParisottoBS15}. What's worse, if the policy has fully converged and its parameters are saturated without plasticity, finetuning on its parameters directly might cause slow improvements and require more training epochs than training from scratch, as found in \cite{DBLP:journals/corr/abs-2306-13812}.

To this end, we propose a partial inheritance method, where the parameters of the policy selected from the previous DP-round are randomly corrupted by fusing with randomly initialized parameters, formulated as:
\begin{equation}\label{eq:fuse}
\theta_{f}^m(\alpha) = \alpha \cdot \theta_\text{best}^{m-1} + (1-\alpha) \cdot \theta_0,
\end{equation}
where $\theta_{f}^m(\alpha)$ represents the fused model parameters, and $0 \leq \alpha \leq 1$ denotes the fusion ratio. 
By fusing pretrained model parameters with random model parameters, we aim to leverage the accumulated knowledge from previous training to accelerate convergence under new reward functions, while also retaining model plasticity. By this definition, the policy improvement process with a new reward function $R$ with $p$ updates can be formulated as
$\mathcal{I}(R, \theta_{f}^m(\alpha), p)$.

\subsubsection{Bayesian Optimization for Searching Fusing Ratio.}
The key question is how to determine $\alpha$. The simplest approach is to uniformly sample $\alpha$ values from [0, 1], obtaining multiple fused parameters through Eq.(\ref{eq:fuse}), and then validating the performance of each $\alpha$ value through $\mathcal{I}(R, \theta_{f}^m(\alpha), T_\text{max})$. However, this exhaustive method is inefficient. To enhance the efficiency of searching for the optimal fusion ratio and reduce training costs, this paper employs a Bayesian Optimization (BO) method based on Gaussian processes (GP) to search for the optimal fusion ratio.

Specifically, we define the relationship between the RL policy performance score and the fusion ratio $\alpha$ as
\begin{align}
    s(\alpha; \theta_{f}^m(\alpha), T_\text{BO}) = V\left(\mathcal{I}\left(R, \theta_{f}^m(\alpha), T_\text{BO}\right)\right), \label{equ:score_function}
\end{align}
where $T_\text{BO}$ denotes the number of updates before calculating $s(\alpha)$. BO assumes that the RL policy performance score $s$ follows a multivariate normal distribution. 
To this, we use Gaussian processes and data samples $D=\{(\alpha_i, s_i)\}_{i=1}^n$ to construct the posterior distribution of the objective function. 

Specifically, we first select several initial points of the fusion ratio $(\alpha_1, \alpha_2, \ldots, \alpha_i)$ and evaluate the corresponding model performance scores $(s_1, s_2, \ldots, s_i)$ using Eq.(\ref{equ:score_function}). Subsequently, we use the initial fusion ratios and their corresponding performance scores to construct the predictive distribution of the objective function. When selecting new evaluation points in subsequent iterations, the Gaussian process model calculates the Expected Improvement (EI) \cite{DBLP:journals/corr/abs-2201-00272} of the evaluation point and selects the point $\alpha_{i+1}$ that maximizes EI as the new evaluation point. The iteration process is carried out in $J$ rounds.

\textbf{Short-Cut Bayesian Optimization.} 
In theory, a Gaussian Process (GP) with ample data $D=\{(\alpha_i, s_i)\}_{i=1}^n$ can precisely model policy performance and identify the best $\alpha$. However, collecting too many samples is costly. Instead, we leverage the observation that policy performance generally diverges early in training, to apply BO with an early stop with a limited set of ample data, significantly speeding up the process, which we call Short-Cut Bayesian Optimization (SC-BO). Note that, even if $\alpha$ isn't optimal, the dynamic population ensures performance improvements continue.

Formally, the fusion ratio search by SC-BO, denoted as $\alpha_\text{SC-BO}$, w.r.t. to reward function $R$ and policy with parameter $\theta$ is defined as: 
\begin{align}
\label{eq:sc_bo}
    \text{SC-BO}(R, \theta) = \arg \max_\alpha s(\alpha, \theta, T_\text{BO}).
\end{align}

Overall, our policy evolution process given the best policy from the previous DP-round $\theta^{m-1}_\text{best}$ and newly generated reward function $R$ can be formulated as:

\begin{equation}
    \mathcal{I}_\text{evolve}(R, \theta^{m-1}_\text{best}, T_\text{max}) = \mathcal{I}(R, \theta^m_{f}(\alpha_\text{SC-BO}), T_\text{max}).
\end{equation}

\subsection{Reward-Policy Co-Evolution} \label{sec:Reward-Policy Co-Evolution}

In this section, we will further break down the mechanics of our reward-policy co-evolution from the following aspects. \textbf{1) Evaluating the reward function:} In every DP-round, we identify the best new reward function by comparing it with the policy that won the last DP-round. This policy's reward is what we use to prompt the LLM for new rewards. The connection between the new reward and the existing policy is closer than with a brand-new policy, making this evaluation method more efficient and reflective of the new rewards' true performance. 
\textbf{2) Guiding the policy improvement:} The best reward from the previous round is then expanded upon by the LLM, generating a new set of reward functions. These new rewards are crafted to further refine the policy, nudging it closer to the optimal one.
\textbf{3) Dynamic population as a filter:} Our dynamic population mechanism acts like a sieve for the reward-policy combination. Only those combinations that outshine the previous round's top performer survive to the next round. This ensures that only the better pairs can advance in the co-evolutionary process.
\textbf{4) Efficient fusion policy selection:} Using SC-BO, we develop a policy that seamlessly builds on its predecessor's strengths and adapts well to new reward functions. This approach ensures efficient resource use and smooth progression, enabling continuous improvement of both rewards and policies with the same amount of interactions as Eureka.

\section{Experiments} \label{sec:Experiments}

We conducted experimental evaluations of the proposed method within the Isaac Gym \cite{makoviychuk2021isaac} RL benchmark and performed comparative analyses against the sparse reward method, human-designed reward methods, and traditional LLM-designed reward function methods.

\subsection{Experiment Settings} \label{sec:Experiment Settings}
\subsubsection{Environments and Tasks} \label{sec:Environments and Tasks}

\begin{figure}[t]
    \centering
    \includegraphics[width=0.98\columnwidth, trim=270 150 270 130, clip]{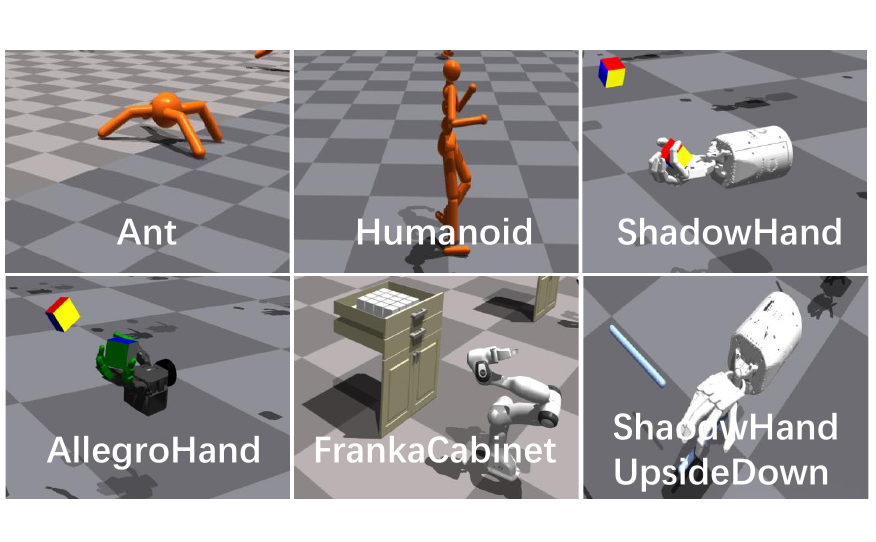}
    \caption{Illustrations of the six robot tasks in our experiment: Ant, Humanoid, ShadowHand, AllegroHand, FrankaCabinet, and ShadowHandUpsideDown.}
    \label{fig:environment}
\end{figure}

We validated our approach on six robotic tasks within Isaac Gym, including Ant, Humanoid, ShadowHand, AllegroHand, FrankaCabinet, and ShadowHandUpsideDown as shown in Fig. \ref{fig:environment}.

\begin{table*}[t]
\centering
\begin{tabular}{ccccccc}
\toprule
Methods& \multicolumn{6}{c}{Task}\\
\cmidrule(lr{0.3em}){2-7}
 -& Ant& Humanoid& ShadowHand& AllegroHand& FrankaCabinet&ShadowHand-U\\
\midrule
Sparse& 6.59±1.44& 5.12±0.49& 0.06±0.039& 0.06±0.02& 0.0007±0.001&0.13±0.09\\
 Human& 10.35±0.12& 6.93±1.38& 6.00±1.02& 11.57±0.53& 0.10±0.05&14.86±6.36\\
 Eureka& 10.25±1.31& 7.24±0.64& 9.56±2.17& 14.60±4.14& 0.31±0.22&8.35±4.35\\
ROSKA-U& \textbf{12.52±1.03}& 8.84±0.69& 24.07±1.80& \textbf{26.80±2.17}& 0.81±0.21&\textbf{23.72±4.96}\\
ROSKA& 12.07±0.60& \textbf{9.10±1.06}& \textbf{24.34±2.84}& 23.22±2.37& \textbf{0.85±0.19}&21.82±5.87\\
\bottomrule
\end{tabular}
\caption{MTS comparison across six tasks, presented as mean ± standard deviation of returns. Our method (ROSKA) consistently achieves superior performance across all tasks, outperforming other methods.}
\label{tab:maintable}
\end{table*}

In our experiments, we employed the large language model GPT-4o to generate reward functions. Testing revealed that this model outperformed GPT-4 on most tasks (with the exception of the FrankaCabinet task) in terms of average performance. The RL method used to validate our proposed approach was Proximal Policy Optimization (PPO) \cite{DBLP:journals/corr/SchulmanWDRK17}. In all experimental methods, the LLM conducted a total of $N=5$ rounds of reward design for each robotic task, generating $K = 6$ reward functions for each round. As a reminder, a DP round could be viewed as the combination of several fixed-sized rounds, and in this section, we use the number of fixed-sized rounds for a clear and fair comparison. For the algorithm proposed in this paper, each reward function underwent policy evolution, where the Gaussian Process was initialized with fusion ratio points $\alpha_\text{initial} = [0.0, 0.2, 0.4, 0.6, 0.8, 1.0]$. Each reward function underwent a total of $J=12$ policy model evaluations, with $T_\text{BO}=200$ for each evaluation. The other settings of our experiments can be found in the Appendix.

\subsubsection{Baseline Methods} \label{sec:Baseline Methods}

To evaluate the performance of ROSKA, we mainly compare it with three baselines, i.e., sparse reward, human reward, and Eureka.\\
\textbf{Sparse Reward (SR)}: SR refers to reward settings that express the task objective. These sparse reward functions are specifically defined in the settings described by Eureka \cite{DBLP:conf/iclr/MaLWHBJZFA24}.\\
\textbf{Human Reward (HR)}: These reward functions are meticulously designed by researchers based on experience. Compared to sparse rewards, these rewards are more refined. Detailed definitions can be found in the environment settings within Isaac Gym \cite{makoviychuk2021isaac}.\\
\textbf{Eureka}: Eureka is a state-of-the-art algorithm that automatically generates reward functions using LLM. It has shown outstanding performance across various robotic tasks. \\
\textbf{ROSKA}: ROSKA is the proposed reward-policy co-evolution framework.\\
\textbf{ROSKA with} Uniform Search (ROSKA-U): To evaluate the effectiveness of the policy evolution with SC-BO, we compared it with a uniform search method. While this method identifies a reasonably suitable fusion ratio, it requires an excessively large training sample size. For more details, see the Appendix.

\subsubsection{Evaluation Metrics} \label{sec:Evaluation Metrics}

We employ Max Training Success (\textbf{MTS}) and Human Normalized Score (\textbf{HNS}) as the primary evaluation metrics in our experiments, consistent with Eureka \cite{DBLP:conf/iclr/MaLWHBJZFA24}. MTS reflects the average value of sparse rewards obtained during training, serving as a key indicator of model performance. HNS measures the algorithm's performance relative to human-designed reward functions. Given the scale differences in sparse reward metrics across tasks, we use the Human Normalized Score to facilitate performance comparison across different methods. 
The Human Normalized Score is calculated as follows:
\begin{equation}
HNS = \frac{MTS_{method} - MTS_{Sparse}}{|MTS_{Human} - MTS_{Sparse}|},\
\label{eq:hns}
\end{equation}
where "Method," "Sparse," and "Human" represent the MTS values obtained from the method under evaluation, the sparse reward method, and the human reward method, respectively.
In addition, we also used Total Training Samples (\textbf{TTS}) to evaluate the sample efficiency of each method. 

\begin{figure}[t] 
    \centering 
    \includegraphics[width=1.\columnwidth,trim=5 10 10 5, clip]{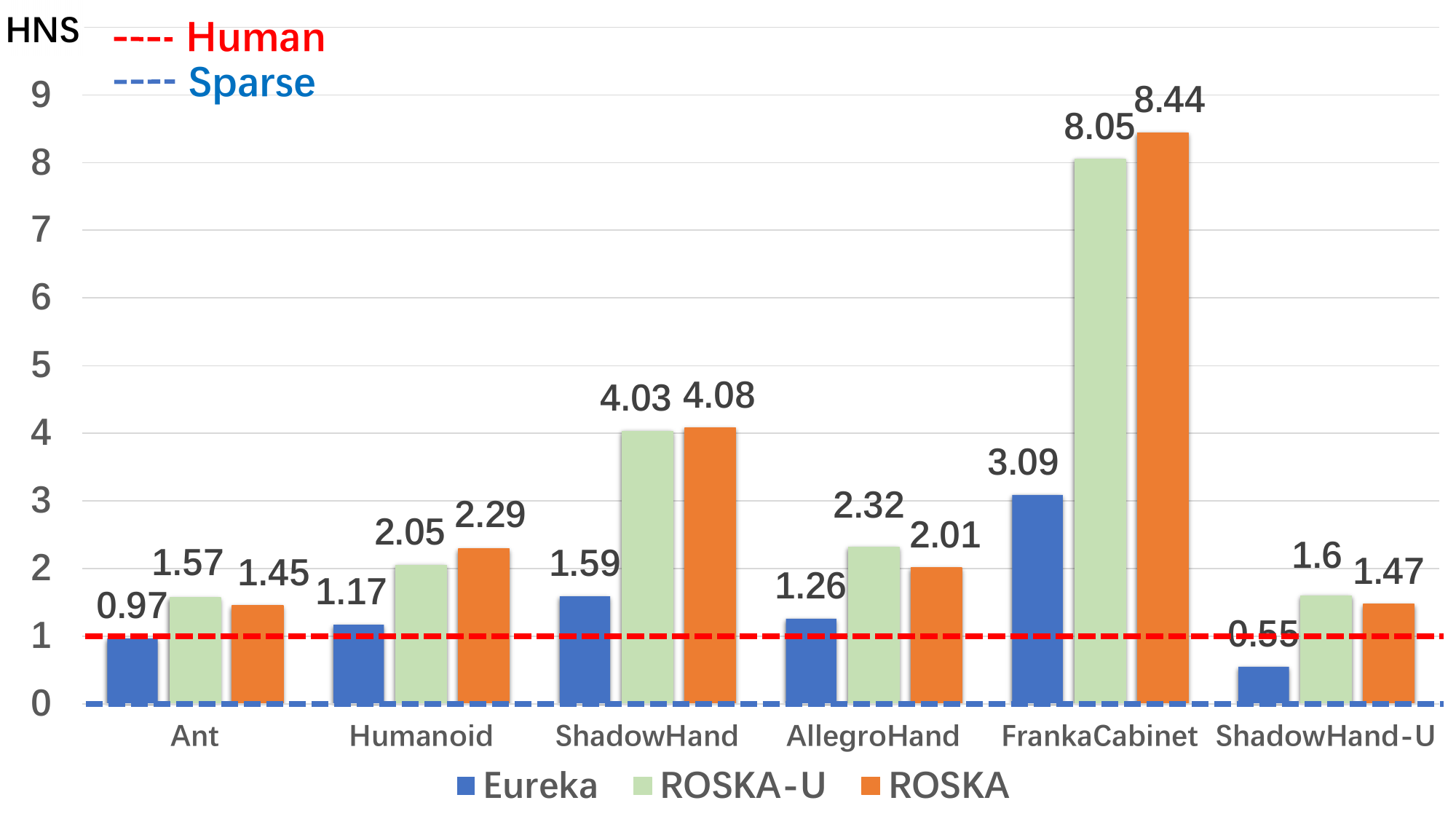} 
    
    \caption{HNS comparison across six robotic tasks, demonstrating that our method consistently outperforms other methods, with substantial improvements across all tasks. }  
    \label{fig:mainfigure} 
\end{figure}

\subsection{Experimental Results and Analysis} \label{sec:Experimental Results and Analysis}
\subsubsection{Comparison to Baseline Methods} \label{sec:Comparison to Baseline Methods}

We compared the MTS of the baselines and our method across six robotic tasks. As shown in Tab \ref{tab:maintable}, our proposed method outperforms all baseline methods in all tasks. For instance, in the ShadowHand task, our method achieved a $154.6\%$ improvement over the Eureka method, and in the ShadowHandUpsideDown task, it achieved a $184.07\%$ improvement.
The Eureka algorithm framework designs reward functions iteratively, with each round geared towards training from scratch using traditional RL methods. This approach cannot guarantee that each round of training will yield a higher score, as the reward functions designed by the LLM might be worse than those from previous rounds. In contrast, ROSKA inherits pretrained policy, effectively ensures an overall positive optimization trend. With the same number of reward design rounds, our method significantly outperforms the Eureka algorithm. For example, in the AllegroHand task, our method outperforms Eureka by $83.56\%$.
Compared to the ROSKA-U method, which is based on a uniform search for $\alpha$, ROSKA achieved better results in Ant, AllegroHand, and ShadowHand-U tasks, while in other tasks, ROSKA-U performed slightly better. However, the training sample size for ROSKA-U is nearly $2.5$ times that of ROSKA, which is further analyzed in ablation studies. This indicates that the ROSKA method can achieve performance comparable to that of the ROSKA-U method while using fewer samples.

We further compared the performance of each algorithm using the Human Normalized Score (HNS) metric, which more intuitively demonstrates the performance of our method relative to human-designed reward functions. HNS=$1$ indicates that the algorithm's performance is equivalent to that of human-designed rewards, and a higher HNS value signifies better algorithm performance. As shown in Fig. \ref{fig:mainfigure}, our method surpasses the performance of expert-designed rewards in all six robotic tasks. Notably, in the ShadowHand and FrankaCabinet tasks, our method exceeds the performance of human-designed rewards by $4$ times and $8$ times, respectively, which is an extraordinary improvement. Compared to Eureka, our method achieved an average improvement of $95.3\%$ on this metric.

\begin{figure}[t] 
    \centering 
    \includegraphics[width=1.\columnwidth,trim=60 30 50 30, clip]{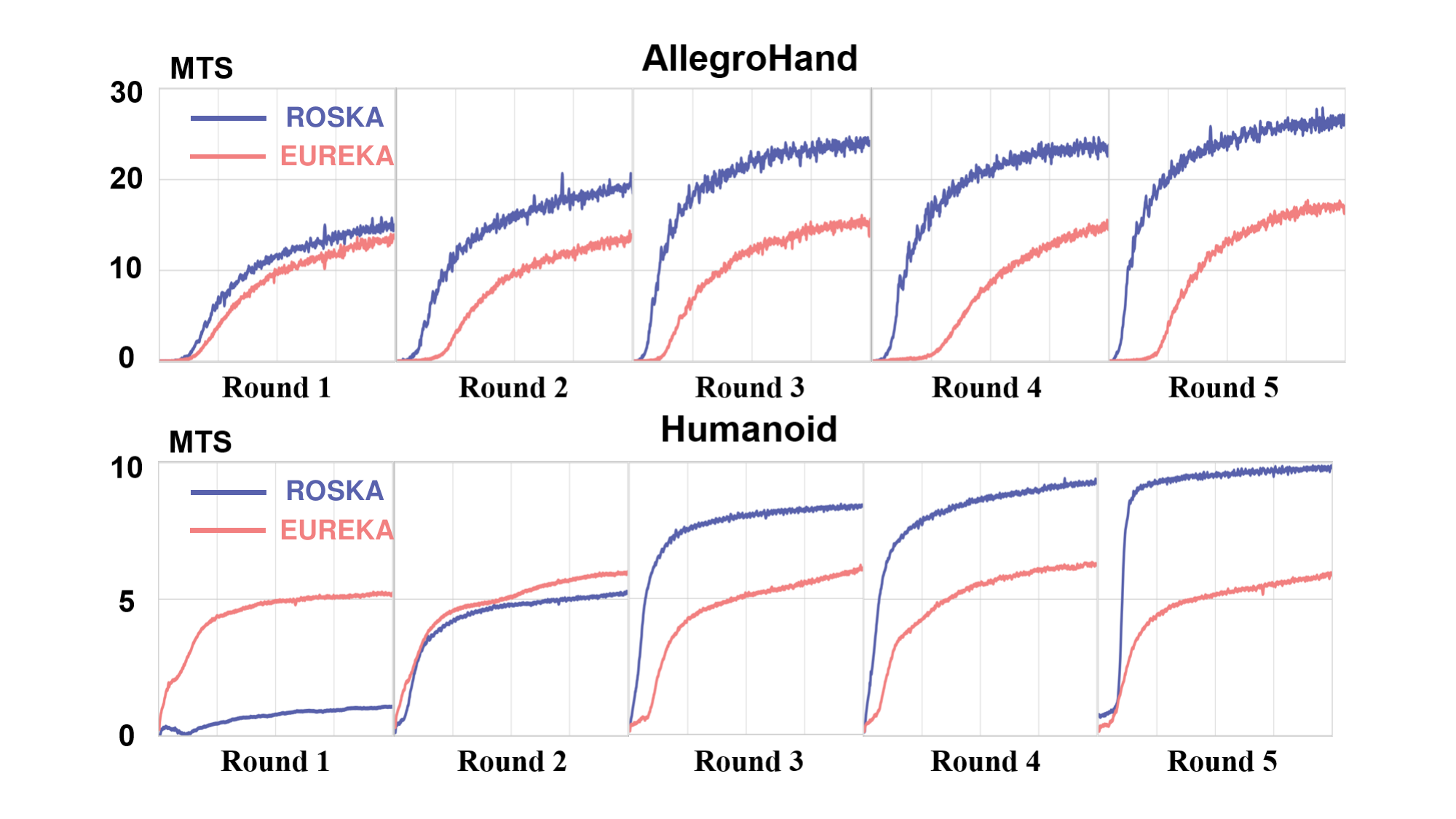} 
    \caption{MTS comparison showing our method's steady improvement and higher scores over rounds, while Eureka struggles with stability. For details, see the appendix.}
    \label{fig:training_curve} 
\end{figure}

To illustrate the reward-policy co-evolution mechanism, we visualized the training curves of ROSKA and Eureka across five rounds of reward function design, as shown in Fig. \ref{fig:training_curve}. Results from the AllegroHand and Humanoid tasks demonstrate that, from the second round onward, ROSKA converges faster and achieves superior performance. This suggests that ROSKA effectively leverages pre-trained policy knowledge to enhance learning efficiency under new reward functions. In contrast, Eureka relies solely on reward function evolution, requires the policy to learn from scratch each round, leading to slower improvement. Notably, even when the LLM-generated reward in the first round of the Humanoid task was suboptimal, subsequent rounds saw rapid policy improvement, further validating the effectiveness of the reward-policy co-evolution mechanism.

\subsubsection{Ablation Studies}\label{sec:ablation}

The ablation study focuses on two key questions: First, the effect of the inherited pretrained model proportion $\alpha$ on policy evolution during reward design
Second, the impact of training sample size on the final policy performance in the proposed Reward-Policy co-evolution method, evaluating the sample efficiency of our approach.

\begin{table}[t]
\centering
\resizebox{0.95\columnwidth}{!}{
\begin{tabular}{cccc}
\toprule
Method& \multicolumn{3}{c}{Task}\\
\cmidrule(lr){2-4}
 -& Ant& Humanoid& ShadowHand\\
\midrule
ROSKA-0\%& 10.25±1.31& 7.24±0.64& 9.56±2.17\\
 ROSKA-50\%& 11.00±0.71& 8.13±1.47&15.90±2.25\\
 ROSKA-100\%& 11.06±1.13& 6.24±3.76&21.58±5.10\\
ROSKA& \textbf{12.07±0.60}& \textbf{9.10±1.06}& \textbf{24.34±2.84}\\
\bottomrule
\end{tabular}
}
\caption{MTS Comparison of the SC-BO method with fixed fusion ratios, showing that SC-BO search method achieves the best performance across tasks.}
\label{tab:fix_alpha}
\end{table}

\textbf{Effectiveness of SC-BO Search for Fusion Ratio.} To evaluate the impact of inheriting the pretrained policy on the final model performance, we conducted experiments comparing the BO search method proposed in this paper with fixed fusion ratios. Fixed fusion ratio refers to a constant $\alpha$ value in each round during the design of rewards, representing the proportion of the historically optimal policy inherited. In our experiments, we selected $\alpha = [0\%, 50\%, 100\%]$ for three sets of experiments, denoted as ROSKA-$0\%$, ROSKA-$50\%$, and ROSKA-$100\%$.
When $\alpha = 0\%$, meaning that each round uses a randomly initialized policy for training. In Tab. \ref{tab:fix_alpha}, ROSKA-$0\%$ underperforms other methods that inherit pretrained parameters in the Ant and ShadowHand tasks, indicating that it's beneficial to inherit pretrained knowledge.
Overall, ROSKA performed the best across all tasks, showing at least a $9\%$ improvement over other methods with fixed fusion ratios. This result demonstrates that a reasonable inheritance ratio can significantly enhance the final performance of the policy.

\begin{table}[t]
\centering
\resizebox{0.98\columnwidth}{!}{
\begin{tabular}{ccccc}
\toprule
Method &TTS&\multicolumn{3}{c}{Task}\\
\cmidrule(lr){3-5}
-&- &Ant& Humanoid& ShadowHand\\
\midrule
Eureka&1 &10.2±1.3&  7.2±0.6& 9.5±2.1\\
\midrule
 ROSKA-U&2.2 &12.5±1.0& 8.8±0.6&24.0±1.8\\
\midrule
  &0.56 &10.7±1.4& 7.2±1.4&13.7±3.0\\
 ROSKA&0.74 &10.9±0.7& 7.1±1.2&14.4±3.7\\
&0.89 &\textbf{12.0±0.6}&  \textbf{9.1±1.0}& \textbf{24.3±2.8}\\
\bottomrule
\end{tabular}
}
\caption{MTS across tasks with varying TTS, presented as mean ± standard deviation. The Eureka method's TTS is used as the baseline (TTS = 1), with other methods' TTS expressed as proportions. }
\label{tab:sample_num}
\end{table}

\textbf{Impact of Training Sample Size on Final Policy Performance.} To validate the Sample Efficiency of the SC-BO method proposed in this paper, we evaluated the results obtained using different sample sizes. As shown in Tab. \ref{tab:sample_num}, even as the sample size decreases, our method continues to achieve good results. Notably, when using only $56\%$ of the sample size required by the Eureka method, our approach still yields competitive results.
For the uniform search method, although it can achieve good results, the required training sample size is extremely large. Our SC-BO-based method achieves comparable performance while using only $40\%$ of the samples required by the uniform search method, significantly reducing the training cost. This indicates that the SC-BO method can efficiently find an optimal fusion ratio, enabling effective inheritance of pretrained policy knowledge and promoting rapid convergence of the policy under new reward functions.

\section{Conclusion}

In conclusion, our integration of large language models (LLMs) with reinforcement learning (RL) through the ROSKA framework has marked a significant leap in resolving language-instructed robotic tasks. ROSKA transcends traditional RL by enabling a co-evolution of reward functions and policies, synchronized to enhance each other. This symbiotic advancement optimizes the efficiency and effectiveness of robotic learning from language instructions. Extensive experiments witnessed an average improvement of 95.3\% across complex robotic tasks with fewer samples, confirming the framework's potency. It highlights the capability of ROSKA to bolster robotic adaptability and autonomy, advancing the frontier of autonomous robotics. 

\section{Acknowledgements}

This work is supported in part by the National Natural Science Funds for Distinguished Young Scholar under Grant 62325307, in part by the National Natural Science Foundation of China under Grants 6240020443, 62073225, 62203134, in part by the Natural Science Foundation of Guangdong Province under Grants 2023B1515120038, in part by Shenzhen Science and Technology Innovation Commission (20231122104038002, 20220809141216003, KJZD20230923113801004), in part by the Guangdong “Pearl River Talent Recruitment Program” under Grant 2019ZT08X603, in part by the Guangdong “Pearl River Talent Plan” under Grant 2019JC01X235, in part by the Scientific Instrument Developing Project of Shenzhen University under Grant 2023YQ019, in part by the Major Key Project of PCL (No. PCL2024A04, No. PCL2023AS203).

\bibliography{aaai25}
\clearpage 
\section{Appendix}

\subsection{Algorithm Description}

In this section we will introduce the algorithm flow of ROSKA, as illustrated in Alg. \ref{alg:ROSKA}. Similar Eureka, our ROSKA employs a multi-round iterative process to design reward functions. In the first round, the large language model (LLM) GPT-4o generates $K=6$ reward functions based on the provided initial prompt information in a zero-shot manner. The ``initial prompt" will be shown in the following section. These reward functions are then trained using the PPO reinforcement learning algorithm for $500$ epochs. Then, a selection process is conducted, and the reward function with the best-performed undergoes an additional $2500$ epochs of training to obtain the pretrained policy $\theta$ of first round. This pretrained model and its associated reward function are considered as the best-performed policy $\theta^{1}_{best}$ and reward function $R^{1}_\text{DP,best}$, respectively.

\begin{algorithm}[ht] 
\caption{Reward-Policy Co-Evolution}
\label{alg:ROSKA}
\begin{algorithmic}[1]
\REQUIRE LLM, task description $I_d$, environment code $I_e$, rounds $N$, iterations $J$, batch size $K$, initial policy $\theta_0$
\STATE \textcolor{deepblue}{// First round reward functions generation}
\STATE  $R_\text{DP}^1 = \text{LLM}(I_d, I_e)$
\STATE \textcolor{deepblue}{// Reward evolution by Eq. (\ref{equ:reward_evolution})} 
\STATE Obtain $\theta^1_\text{best}$ and $R^1_\text{Dp,best}$
\FOR{$m = 2$ to $N$}
\STATE \textcolor{deepblue}{// Reward functions generation}
    \STATE $\mathbf{R}_\text{DP}^{m} = \text{LLM}(I_d, I_e, R_\text{DP, best}^{m-1}, V(\theta^{m-1}_\text{best}))$
    \FOR{$R^m_k \in \mathbf{R}_\text{DP}$}
         \STATE \textcolor{deepblue}{// Policy evolution} 
         \STATE $\alpha_\text{SC-BO}= \arg \max_\alpha s(\alpha, \theta, T_\text{BO})$
         \STATE $\theta_{f}^m(\alpha_\text{SC-BO}) = \alpha_\text{SC-BO} \cdot \theta_\text{best}^{m-1} + (1-\alpha_\text{SC-BO}) \cdot \theta_0$
         \STATE $\mathcal{I}_\text{evolve}(R, \theta^{m-1}_\text{best}, T_\text{max}) = \mathcal{I}(R, \theta^m_{f}(\alpha_\text{SC-BO}), T_\text{max})$
    \ENDFOR
    \IF{$\theta^m_f(\alpha_{\text{SC-BO}})$ outperforms $\theta^{m-1}_{\text{best}}$}
    \STATE \textcolor{deepblue}{// Reward evolution}
        \STATE Update $\theta^{m}_{best}$ and $R^m_{DP, best}$
    \ENDIF
\ENDFOR
\STATE \textbf{Output:} Best-performed policy $\theta_{\text{best}}$ 
\end{algorithmic}
\end{algorithm}

In the subsequent iterative rounds, the LLM generates a new set of $K=6$ reward functions based on the feedback prompt, which contains the information of existing ``best-performed reward function" and the ``measurement of the best-performed policy". We then apply the SC-BO algorithm to iteratively search for the fusion ratio $\alpha_\text{SC-BO}$ (as defined in Eq. \ref{eq:sc_bo}), with $J=12$ iterations per reward function, each iteration consumes $T_\text{BO}=200$ training epochs. This process yields $6$ policies, each of which undergoes an additional $300$ training epochs. The best-performed policy is selected for an extended training of $2500$ epochs.

If the completed trained policy in this round outperforms the previous best-performed policy, $\theta_\text{DP, best}$ and $R_\text{DP,best}$ will be updated and used as input for the next reward function design round. After completing all the rounds, the model with the highest score is identified as the final policy of ROSKA.


\subsection{Experimental Setup}
In this section, we present the experimental setup for ROSKA-U, provide a description of the IsaacGym tasks, outline the computing infrastructure, and detail the calculation of TTS. The ROSKA code will be released at the following URL: \url{https://github.com/NextMyLove/ROSKA}. 

\textbf{ROSKA based on Uniform Search (ROSKA-U)} To assess the effectiveness of the BO-based policy evolution described, we compare it with a uniform search method. Here, the fusion ratio $\alpha$ is uniformly sampled within the range $[0, 1]$, with $11$ fusion ratios tested. Each fusion policy is trained with $3000$ epochs, and the policy with the best performance is retained. While this method identifies a reasonably suitable fusion ratio, it requires an excessively large training sample size.

\textbf{IsaacGym Tasks}
A brief description of the objectives for these robotic reinforcement learning tasks is as follows:
\begin{itemize}
    \item \textbf{Ant:} Train the ant robot to run forward as fast as possible.
    \item \textbf{Humanoid:} Train the humanoid robot to run as fast as possible.
    \item \textbf{ShadowHand:} Train the ShadowHand to spin an object to a target orientation.
    \item \textbf{AllegroHand:} Train the AllegroHand to spin an object to a target orientation.
    \item \textbf{FrankaCabinet:} Enable the Franka robot to open a cabinet door.
    \item \textbf{ShadowHandUpsideDown (ShadowHand-U):} Train the ShadowHand, with its palm facing down, to spin an object to a target orientation.
\end{itemize}
More details are shown in Tab. (\ref{isaacgym}).

\begin{table*}[ht]
\centering
\resizebox{\textwidth}{!}{
\begin{tabular}{|>{\centering\arraybackslash}m{4cm}|>{\centering\arraybackslash}m{6cm}|>{\centering\arraybackslash}m{5cm}|}
\hline
\multicolumn{3}{|c|}{\textbf{IsaacGym Environments}} \\ \hline
\textbf{Environment (observation dim, action dim)} & \textbf{Task description} & \textbf{Task fitness function $F$} \\ \hline
FrankaCabinet (23, 9) & To open the cabinet door & $1[\text{cabinet\_pos} > 0.39]$ \\ \hline
Ant (60, 8) & To make the ant run forward as fast as possible & $\text{cur\_dist} - \text{prev\_dist}$ \\ \hline
AllegroHand (88, 16) & To make the hand spin the object to a target orientation & number of consecutive successes where current success is $1[rot\_dist < 0.1]$\\ \hline
Humanoid (108, 21) & To make the humanoid run as fast as possible & $\text{cur\_dist} - \text{prev\_dist}$\\ \hline 
ShadowHand-U (211, 20) & To make the shadow hand, whose palm faces down, spin the object to a target orientation & number of consecutive successes where current success is $1[\text{rot\_dist} \leq \text{success\_tolerance}]$\\ \hline
ShadowHand (211, 20) & To make the shadow hand spin the object to a target orientation & number of consecutive successes where current success is $1[\text{rot\_dist} < 0.1]$ \\ \hline
\end{tabular}
}
\caption{Description of various tasks in the IsaacGym environments, including observation and action dimensions, task descriptions, and the corresponding task fitness functions. We use the mean sparse reward across all agents as a proxy for return. This approach is justified because the average sparse reward effectively captures the collective performance of the agents. Additionally, sparse rewards are often aligned with long-term objectives in such environments, making the mean reward a reasonable and practical approximation of the return for evaluating the policy's performance.}
\label{isaacgym}
\end{table*}

%

\textbf{Computing Infrastructure}
The experiments in this paper were conducted on a computer system running Ubuntu 22.04.4 LTS, equipped with two 4090 GPUs. Since the training time required for training on different tasks is different, for intuitive comparison, we use the TTS for comparison.


\textbf{Calculation of TTS:}
We detail the calculation of the Training Sample Size (TTS) ratio for different methods discussed in the paper.
We use "epochs" to measure the sample size for each method, ensuring that each epoch contains the same number of transitions across all methods.

\textbf{Eureka:}  The reward function are generated for $5$ iterative rounds, with each round generating $6$ reward functions. In Eureka, each reward function is trained for $3000$ epochs. The total training epochs for Eureka are calculated as: $5 (\text{rounds}) \times 6 (\text{reward functions}) \times 3000 (\text{epochs}) = 90000 (\text{epochs})$.
We use the TTS of the Eureka method as the baseline and compare the TTS of other methods to it, that is, TTS ratio = Method(TTS)/Eureka(TTS).

\textbf{ROSKA:}  
The ROSKA method is divided into two parts: the first round and subsequent rounds. \textbf{First Round:} $500 (\text{epochs}) \times 6 (\text{reward functions}) + 2500 (\text{epochs}) = 5500 
 (\text{epochs})$. \textbf{Subsequent Rounds:} $4 (\text{rounds}) \times ( 6 (\text{reward\, functions}) \times 200 (\text{epochs}) \times 12 (\text{iterations}) + 300 (\text{epochs}) \times 6 (\text{reward\, functions}) + 2500 (\text{epochs}) ) = 74800  (\text{epochs})$. Thus, the total training epochs for the ROSKA is: $5500 (\text{epochs}) + 74800 (\text{epochs})= 80300 (\text{epochs})$, and the TTS ratio of ROSKA is:
\[
 \text{TTS} = \frac{80300 (\text{epochs})}{90000 (\text{epochs})} \approx 0.89.
\]

\textbf{ROSKA-U:}  
The ROSKA-U method, based on uniform search, also involves the first and subsequent rounds. \textbf{First Round:} $6 (\text{rounds}) \times 3000 (\text{epochs}) = 18000 \text{epochs}$.
\textbf{Subsequent Rounds:} $4 (\text{rounds}) \times 6 (\text{reward functions}) \times (500 (\text{epochs}) \times 11(\text{uniform search}) + 2500(\text{epochs})) = 210000 \text{epochs}$.
The TTS ratio of ROSKA-U is:
\[
\text{TTS} = \frac{210000 (\text{epochs})}{90000 (\text{epochs})} \approx 2.2.
\]

\textbf{ROSKA-0.74:}  
ROSKA(0.74) is a variant of ROSKA, with changes in the number of SC-BO iterations and training epochs for the best reward-policy combination. \textbf{First Round:} $200 (\text{epochs}) \times 6(\text{reward functions}) + 2800 (\text{epochs}) = 4000 \text{epochs}$. \textbf{Subsequent Rounds:} $4 (\text{rounds}) \times (6 (\text{reward functions}) \times 200 (\text{epochs}) \times 12 (\text{iterations}) + 1300 (\text{epochs})) = 62800 \text{epochs}$. Thus, the total training epochs is: $\text{TTS} = 4000(\text{epochs}) + 62800(\text{epochs}) = 66800 \text{(epochs)}$.
Thus, the TTS ratio of ROSKA-0.74 is:
\[
\text{TTS} = \frac{66800(\text{epochs})}{90000(\text{epochs})} \approx 0.74.
\]

\textbf{ROSKA-0.56:}  
ROSKA-0.56 is another variant of ROSKA, differing in SC-BO iterations and training epochs for the best reward-policy combination. \textbf{First Round:} $200 (\text{epochs}) \times 6 (\text{reward functions}) + 2800 (\text{epochs}) = 4000 
 (\text{epochs})$. \textbf{Subsequent Rounds:}  $4 (\text{rounds}) \times 6 (\text{reward functions} \times 200 (\text{epochs}) \times 9 (\text{iterations}) + 800 (\text{epochs})) = 46400 \text{epochs}$. The total training epochs for ROSKA-0.56 is: 4000 (\text{epochs}) + 46400 (\text{epochs}) = 50400 \text{epochs}
The TTS ratio for ROSKA(0.56) relative to Eureka is:
\[
\text{TTS} = \frac{50400(\text{epochs})}{90000 (\text{epochs})} \approx 0.56.
\]

\subsection{Discussion}

\textbf{Discussion on Expanding ROSKA} In this work, we introduced ROSKA, a method designed to co-evolve rewards and policies using Large Language Models (LLMs). We believe that ROSKA provides a robust foundation for scaling to more complex tasks, particularly in the context of long-horizon tasks with hierarchical or compositional reward structures. Our current evaluation of ROSKA focused on fundamental and representative tasks, ensuring that the method's efficacy is established across a broad spectrum of scenarios.

However, the potential of ROSKA extends far beyond these basic tasks. Our ongoing efforts aim to integrate Visual Language Models (VLM) to parse both visual inputs and extensive textual instructions, facilitating curriculum-based learning for tasks with extended horizons. This direction introduces new challenges, including the use of VLM to formulate curricula from visual data and text, assessing the quality of such curricula, and selecting autonomous learning trajectories from diverse curriculum combinations.

We believe these advancements will significantly enhance the autonomy and sophistication of robotic learning systems. As part of this expansion, we envision ROSKA playing a foundational role in curriculum learning. Specifically, each sub-curriculum could leverage ROSKA to refine its reward function and policies. By transitioning from a singular sparse reward to a composite of sparse rewards that span multiple curricula, ROSKA could evolve into a continuous learner. Additionally, by transforming the dynamic reward population into a dynamic curriculum population, we enable the co-evolution of curricula, rewards, and policies, further enhancing the method's adaptability and scalability in complex learning environments.

We highlight the foundational role of ROSKA in curriculum learning and its potential for future applications in extended horizon tasks. Through this work, we aim to establish a clear connection between foundational reward-policy co-evolution and more complex learning frameworks, paving the way for future explorations in autonomous robotic learning systems.

\textbf{Discussion on the Challenges of LLM in Dynamic Environments} In this work, we leverage Large Language Models as a key component for reasoning task instructions. One of the challenges associated with LLMs is their need for careful tuning and optimization, which may not guarantee optimal performance across all tasks, especially in highly dynamic or unpredictable environments. We fully acknowledge this issue, and we believe that with the continuous emergence of more advanced LLMs and evolving prompting techniques, these challenges will gradually be mitigated. Moreover, we argue that, aside from humans, LLMs currently represent one of the most promising tools for autonomous reasoning and understanding of complex task instructions.

While we agree that the optimization of LLMs is an important consideration, we emphasize that our method is not solely dependent on LLMs for performance. The advancements in LLM technology are orthogonal to the core mechanics of our framework, meaning that improvements in LLMs do not conflict with or undermine the effectiveness of our approach. We view the integration of LLMs as part of a broader trend towards enabling more sophisticated, autonomous systems that can efficiently process and reason about complex tasks in a variety of environments.

In response to concerns about efficiency, particularly in time-sensitive scenarios, we would like to clarify that our approach, similar to many advanced reinforcement learning methods, places primary focus on the training process of skills. During training, LLMs play a key role in the iterative refinement of reward functions and policies, while the deployment of these skills leverages smaller, pre-trained policy networks. As such, the use of LLMs does not impose a significant overhead on time-critical real-world applications, since the trained policies can be directly deployed without the need for constant interaction with the LLMs during execution.

\textbf{Discussion on Evolution of total reward computation}

we prepare a detailed illustration of the rewards obtained after each round of evolution in the following.

\begin{table}[ht]
    \centering
    \renewcommand{\arraystretch}{1.5} 
    \resizebox{1.0\columnwidth}{!}{
        \begin{tabular}{|c|l|}
            \hline
            \textbf{Reward} & \textbf{Formula} \\ \hline
            Total Reward  1 & 
            \begin{tabular}[t]{@{}l@{}}
                \texttt{total\_reward} = $2.0 \times \texttt{forward\_velocity\_rew}$ \\ 
                $+ 1.0 \times \texttt{upright\_rew}$ \\ 
                $+ 1.0 \times \texttt{progress\_rew}$
            \end{tabular} \\ \hline
            Total Reward 2 & 
            \begin{tabular}[t]{@{}l@{}}
                \texttt{total\_reward} = $2.0 \times \texttt{forward\_velocity\_rew}$ \\ 
                $+ 0.5 \times \texttt{upright\_rew}$ \\ 
                $+ 0.5 \times \texttt{progress\_rew}$ \\ 
                $- 0.1 \times \texttt{angular\_velocity\_penalty}$
            \end{tabular} \\ \hline
            Total Reward 3 & 
            \begin{tabular}[t]{@{}l@{}}
                \texttt{total\_reward} = $2.0 \times \texttt{forward\_velocity\_rew}$ \\ 
                $+ 0.8 \times \texttt{upright\_rew}$ \\ 
                $+ 0.5 \times \texttt{progress\_rew}$ \\ 
                $+ 0.5 \times \texttt{torso\_height\_reward}$ \\ 
                $- 0.3 \times \texttt{angular\_velocity\_penalty}$
            \end{tabular} \\ \hline
            Total Reward 4 & 
            \begin{tabular}[t]{@{}l@{}}
                \texttt{total\_reward} = $2.0 \times \texttt{forward\_velocity\_rew}$ \\ 
                $+ 0.8 \times \texttt{upright\_rew}$ \\ 
                $+ 0.5 \times \texttt{progress\_rew}$ \\ 
                $+ 0.5 \times \texttt{torso\_height\_reward}$ \\ 
                $- 0.3 \times \texttt{angular\_velocity\_penalty}$
            \end{tabular} \\ \hline
            Total Reward 5 & 
            \begin{tabular}[t]{@{}l@{}}
                \texttt{total\_reward} = $2.5 \times \texttt{forward\_velocity\_rew}$ \\ 
                $+ 1.5 \times \texttt{upright\_rew}$ \\ 
                $+ 1.0 \times \texttt{progress\_rew}$ \\ 
                $+ \texttt{torso\_height\_reward}$ \\ 
                $+ \texttt{angular\_velocity\_penalty}$
            \end{tabular} \\ \hline
        \end{tabular}
    }
    \caption{Total Reward Computation and component weights of Best Reward Function during Co-evolution.}
    \label{tab:rewards}
\end{table}

After analyzing these rewards, we have identified some potential correlations and characteristics, which are summarized as follows.
Incremental Addition of Rewards and Penalties: Starting from a straightforward reward function that emphasizes speed and upright posture, the reward structure evolved with the addition of torso height rewards, angular velocity penalties, and tilt angle penalties. This incremental approach guided the model progressively to achieve balance, stability, and control over speed simultaneously.
Dynamic Weight Adjustments: At each round, the reward and penalty weights were adjusted based on emerging needs. This dynamic weighting adjust the policy learning toward essential factors, such as posture and stability, while also balancing the importance of forward progress and other rewards. The fine-tuning of weights provided a means to direct the policy's learning focus adaptively as it developed.




\subsection{Limitations}

In our experiments, we found that the reward functions generated from LLM are not always stable. This situation also exists in the other LLM-based reward generation methods. 
Specifically, for the same language prompt, the reward functions generated by LLM multiple times may vary greatly, especially in the first round of zero-shot generation.
This problem is particularly evident in more complex tasks, such as Humanoid task, which may cause all reward functions generated in the first round to be unworkable. 
In this case, we have to query the LLM multiple times to ensure that at least one of the reward functions generated in the first round is effective before proceeding with subsequent iterative training. We also believe that adjusting the initial prompt could yield a more stable reward function in the first round.

We also found that if LLM generates high-quality reward functions in the first few rounds, the subsequent training process can be significantly accelerated.  
This is because ROSKA uses the reward-policy co-evolution manner, which can flexibly inherit the previously pre-trained knowledge to speed up the current training.
If the LLM generates high-quality reward functions in the initial phase, the ROSKA algorithm can quickly extend this advantage throughout the training process. 
Therefore, if LLM generates high-quality reward functions in the initial stage, ROSKA can gain a greater advantage over other methods.
However, despite the poor performance of the initial reward function, ROSKA can still achieve stable performance improvement through reward-policy co-evolution.

\subsection{Prompts Detail}

The ROSKA framework contains two parts of prompts, initial prompt and feedback prompt.

\textbf{Initial Prompt:} 
\begin{mdframed}[backgroundcolor=lightgray!20, linecolor=lightgray, leftmargin=0cm, rightmargin=0cm, innerleftmargin=0.3cm, innerrightmargin=0.3cm]

\color{black}
You are a reward engineer trying to write reward functions to solve reinforcement learning tasks as effectively as possible. Your goal is to write a reward function for the environment that will help the agent learn the task described in text. Your reward function should use useful variables from the environment as inputs. As an example, the reward function signature can be:

\texttt{@torch.jit.script} \\
\texttt{def compute\_reward(object\_pos: torch.Tensor, goal\_pos: torch.Tensor) -> Tuple[torch.Tensor, Dict[str, torch.Tensor]]:} \\
\texttt{...} \\
\texttt{return reward, \{\}}

Since the reward function will be decorated with \texttt{@torch.jit.script}, please make sure that the code is compatible with TorchScript (e.g., use torch tensor instead of numpy array). Make sure any new tensor or variable you introduce is on the same device as the input tensors. The output of the reward function should consist of two items:
\begin{enumerate}
    \item the total reward,
    \item a dictionary of each individual reward component.
\end{enumerate}
The code output should be formatted as a python code string: \texttt{"```python ... ```"}.

Some helpful tips for writing the reward function code:
\begin{enumerate}
    \item You may find it helpful to normalize the reward to a fixed range by applying transformations like \texttt{torch.exp} to the overall reward or its components.
    \item If you choose to transform a reward component, then you must also introduce a temperature parameter inside the transformation function; this parameter must be a named variable in the reward function and it must not be an input variable. Each transformed reward component should have its own temperature variable.
    \item Make sure the type of each input variable is correctly specified; a float input variable should not be specified as \texttt{torch.Tensor}.
    \item Most importantly, the reward code's input variables must contain only attributes of the provided environment class definition (namely, variables that have prefix \texttt{self.}). Under no circumstance can you introduce new input variables.
\end{enumerate}
\end{mdframed}

\textbf{Feedback Prompt:} It consist of the code and the measurement of the best-performed reward function. As shown in the following text, the deepblue text indicates the code of reward function, and the reddishbrown indicates the measurement, respectively.\\
\begin{mdframed}[backgroundcolor=mygray, linecolor=lightgray, leftmargin=0cm, rightmargin=0cm, innerleftmargin=0.3cm, innerrightmargin=0.3cm]
\color{deepblue}
\textbf{Best-performed reward function}
\small 
\begin{verbatim}
@torch.jit.script
def compute_reward(
    object_rot: torch.Tensor,
    goal_rot: torch.Tensor,
    object_angvel: torch.Tensor
) -> Tuple[torch.Tensor,\
Dict[str, torch.Tensor]]:

    # Quaternion conjugate to measure
    # orientation difference
    quat_conj = torch.cat(
        (goal_rot[:, 0:1], 
        -goal_rot[:, 1:]), dim=1
    )
    quat_diff = torch.cat(
        (
            object_rot[:, 0:1] 
            * quat_conj[:, 0:1] 
            - (object_rot[:, 1:] 
            * quat_conj[:, 1:])
            .sum(dim=1, keepdim=True),
            object_rot[:, 0:1] 
            * quat_conj[:, 1:] 
            + quat_conj[:, 0:1] 
            * object_rot[:, 1:] 
            + torch.cross(
                object_rot[:, 1:], 
                quat_conj[:, 1:], dim=1
            )
        ), dim=1
    )
    
    # Calculate orientation difference
    orientation_diff = torch.abs(
        1.0 - quat_diff[:, 0]
    ) + torch.norm(
        quat_diff[:, 1:], 
        p=2, dim=1
    )
    
    # Calculate Angular velocity penalty
    angvel_penalty = torch.norm(
        object_angvel, 
        p=2, dim=1
    )
    
    # Temperature parameters for reward 
    # normalization
    temp_orientation_diff = 0.1  
    temp_angvel_penalty = 0.1    
    temp_orientation_diff_decrease = 0.1  
    
    # Exponential transformations 
    # for each component
    orientation_diff_reward = torch.exp(
        -orientation_diff 
        / temp_orientation_diff
    )
    angvel_penalty_reward = torch.exp(
        -angvel_penalty 
        / temp_angvel_penalty
    )
    orientation_diff_decrease_reward = \
    torch.exp(
        -orientation_diff 
        / temp_orientation_diff_decrease
    )
    
    # Consolidate the reward calculation
    reward = (
        4.0 * orientation_diff_reward 
        - 2.0 * angvel_penalty_reward 
        + 2.0 * \
        orientation_diff_decrease_reward
    )
    
    # Return the total reward 
    # and components
    reward_dict = {
        'orientation_diff_reward': 
        orientation_diff_reward,
        'angvel_penalty_reward': 
        angvel_penalty_reward,
        'orientation_diff\
        _decrease_reward': 
        orientation_diff_decrease_reward
    }
    return reward, reward_dict
\end{verbatim}
\color{reddishbrown}
\textbf{Measurement of the best-performed policy:}\\
We trained a RL policy using the provided reward function code and tracked the values of the individual components in the reward function as well as global policy metrics such as success rates and episode lengths after every 300 epochs and the maximum, mean, minimum values encountered:

\textbf{orientation\_diff\_reward:} ['0.01', '0.18', '0.19', '0.19', '0.20', '0.19', '0.19', '0.20', '0.19', '0.19'], Max: 0.20, Mean: 0.18, Min: 0.01 \\
\textbf{angvel\_penalty\_reward:} ['0.01', '0.01', '0.00', '0.00', '0.00', '0.00', '0.00', '0.00', '0.00', '0.00'], Max: 0.15, Mean: 0.01, Min: 0.00 \\
\textbf{orientation\_diff\_decrease\_reward:} ['0.01', '0.18', '0.19', '0.19', '0.20', '0.19', '0.19', '0.20', '0.19', '0.19'], Max: 0.20, Mean: 0.18, Min: 0.01 \\
\textbf{task\_score:} ['0.00', '3.68', '11.95', '14.86', '16.91', '18.31', '19.75', '19.71', '21.48', '21.64'], Max: 22.61, Mean: 15.85, Min: 0.00 \\
\textbf{episode\_lengths:} ['7.96', '419.39', '508.57', '513.89', '538.24', '548.40', '541.51', '518.13', '546.33', '557.34'], Max: 568.58, Mean: 495.31, Min: 7.96

Please carefully analyze the policy feedback and provide a new, improved reward function that can better solve the task. Some helpful tips for analyzing the policy feedback:
\begin{enumerate}
    \item If the success rates are always near zero, then you must rewrite the entire reward function.
    \item If the values for a certain reward component are near identical throughout, then this means RL is not able to optimize this component as it is written. You may consider:
        \begin{enumerate}
            \item Changing its scale or the value of its temperature parameter
            \item Re-writing the reward component 
            \item Discarding the reward component
        \end{enumerate}
    \item If some reward components' magnitude is significantly larger, then you must re-scale its value to a proper range.
\end{enumerate}

Please analyze each existing reward component in the suggested manner above first, and then write the reward function code. The output of the reward function should consist of two items:
\begin{enumerate}
    \item the total reward,
    \item a dictionary of each individual reward component.
\end{enumerate}

The code output should be formatted as a python code string: \texttt{"```python ... ```"}.

Some helpful tips for writing the reward function code:
\begin{enumerate}
    \item You may find it helpful to normalize the reward to a fixed range by applying transformations like \texttt{torch.exp} to the overall reward or its components.
    \item If you choose to transform a reward component, then you must also introduce a temperature parameter inside the transformation function; this parameter must be a named variable in the reward function and it must not be an input variable. Each transformed reward component should have its own temperature variable.
    \item Make sure the type of each input variable is correctly specified; a float input variable should not be specified as \texttt{torch.Tensor}.
    \item Most importantly, the reward code's input variables must contain only attributes of the provided environment class definition (namely, variables that have prefix \texttt{self.}). Under no circumstance can you introduce new input variables.
\end{enumerate}
\end{mdframed}


\end{document}